\documentclass[11pt]{article}
\usepackage{coling2020}
\usepackage{times}
\usepackage{url}
\usepackage{latexsym}

\usepackage{latexsym}
\usepackage[dvipdfmx]{graphicx}
\usepackage{tikz}
\usepackage{listings}

\usepackage{amsmath,amsthm,amssymb,ascmac}
\usepackage{fancybox}
\usepackage{psfrag}
\usepackage{bm}
\usepackage{wrapfig}
\usepackage{color}
\usepackage{url}
\usepackage{multirow}
\usepackage{booktabs}
\usepackage{subfigure}
\usepackage{wrapfig}

\makeatletter
\newcommand{\figcaption}[1]{\def\@captype{figure}\caption{#1}}
\newcommand{\tblcaption}[1]{\def\@captype{table}\caption{#1}}
\makeatother

\newcommand{\fujisyo}[1]{#1}

\title{
Pointing to Subwords for Generating Function Names in Source Code} 
\author{Shogo Fujita$^1$, Hidetaka Kamigaito$^1$, Hiroya Takamura$^1$$^,$$^2$ and Manabu Okumura$^1$ \\
  $^1$Tokyo Institute of Technology \\
  $^2$National Institute of Advanced Industrial Science and Technology (AIST) \\
  {\tt \{fujisyo,kamigaito,oku\}@lr.pi.titech.ac.jp} \\
   \tt takamura@pi.titech.ac.jp
  }

\date{}

\begin{document}
\maketitle
\begin{abstract}
We tackle the task of automatically generating a function name from source code.
Existing generators face difficulties in generating low-frequency or out-of-vocabulary subwords.
In this paper, we propose two strategies for copying low-frequency or out-of-vocabulary subwords in inputs.
Our best performing model showed an improvement over the conventional method in terms of our modified F1 and accuracy on the Java-small and Java-large datasets.
\end{abstract}

\section{Introduction}

\begin{figure}[h]
\centering
\includegraphics[width=6cm]{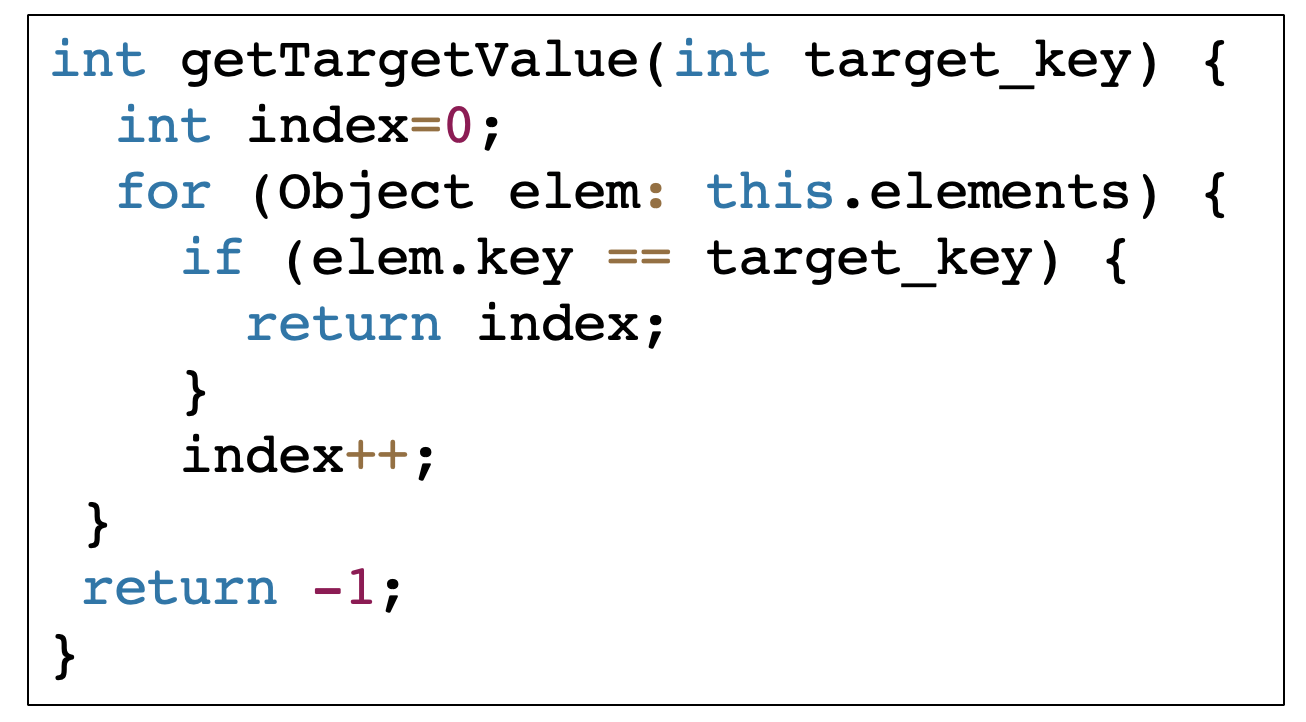}
\caption{Example of a function and its inappropriate name.~(Java)}\label{figinit} 
\end{figure}

Programmers often share source code on sharing services such as GitHub.\footnote{https://github.com} 
Since they can freely define function names in the source code, 
the names are not necessarily reminiscent of the actual behavior of the functions.
For example, the function in Figure~\ref{figinit} returns the index of {\ttfamily elem} whose {\ttfamily elem.key} is the same as {\ttfamily target\_key}. 
However, the function name would be inappropriate as it implies that the function returns the value of the object.
Such a function name adversely affects readability and sometimes causes bugs, especially in collaborative environments.
A proper function name such as {\ttfamily indexOfTarget} in this case, instead of {\ttfamily getTargetValue}, can help programmers understand the code efficiently and avoid possible bugs~\cite{Takang1996TheEO,Binkley2013}.
Automatically generating such function names has been studied as a generation task in natural language processing~\cite{iyer-etal-2016-summarizing}.

\blfootnote{
    \hspace{-0.65cm}  
    This work is licensed under a Creative Commons 
    Attribution 4.0 International Licence.
    Licence details:
    \url{http://creativecommons.org/licenses/by/4.0/}.
}

Recently, various neural network-based approaches have been proposed to solve this problem by generating a function name from given source code
\cite{pmlr-v48-allamanis16,Alon:2018:GPR:3296979.3192412,fernandes2018structured}.
In these approaches, a function name is treated as a sequence of subwords (\texttt{get}, \texttt{Target} and \texttt{Value} in Figure 1).
Since these approaches heavily rely on a subword-based predefined dictionary to generate a function name, it is difficult to generate a function name containing low-frequency or unknown subwords. 

To solve this problem, we propose a method for outputting low-frequency or unknown words using a copy mechanism corresponding to a tree structure, and a method for replacing a specific word with a special token.
We extend code2seq~\cite{alon2018codeseq} by using these methods. Code2seq 
converts source code into an tree-structured representation, called Abstract Syntax Tree (AST), before encoding.
The input for the encoder is not just a sequence of tokens but a set of paths from a leaf to another leaf in the tree.
Thus, the existing copy mechanisms~\cite{gulcehre-etal-2016-pointing,gu-etal-2016-incorporating,Yang2018/05,hsu-etal-2018-unified,cohan-etal-2018-discourse} cannot be directly applied.

We observed that our best-performing model was the one that uses a combination of a hierarchical copy mechanism and a strategy to replace the most frequent word in an input snippet of source code with a delexicalized placeholder.
In particular, the score of the best-performing model was increased in terms of our modified F1 and accuracy, calculated on the Java-small and Java-large\footnote{https://github.com/tech-Srl/code2seq\#datasets} datasets by \newcite{alon2018codeseq}.

\begin{figure}[t]
\centering
\includegraphics[width=0.8\columnwidth]{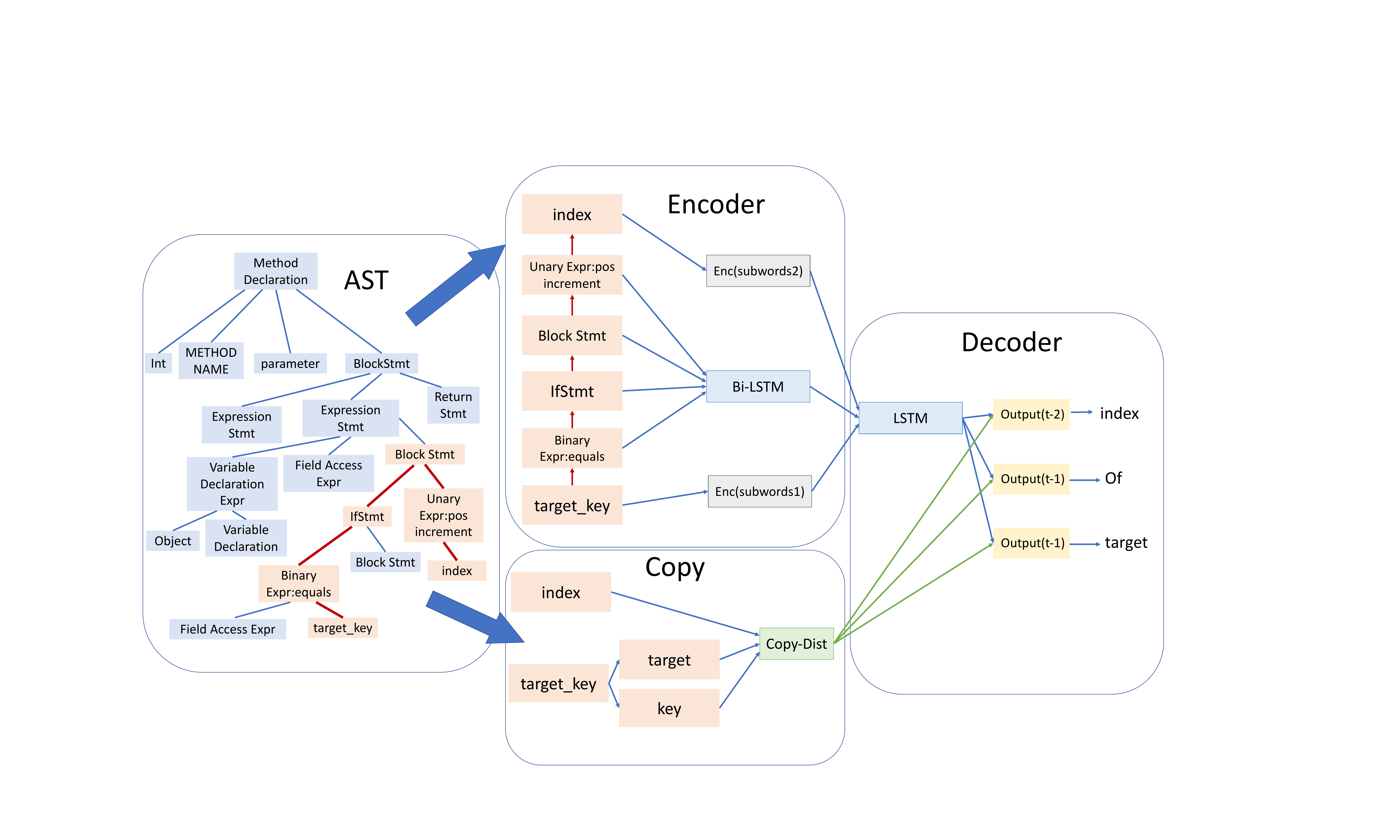}
\caption{
Overview of the function naming with our model.
}\label{ourmodel} 
\end{figure}

\section{Code2seq}
\label{sec:code2seq}
We first describe code2seq~\cite{alon2018codeseq}, an existing model that we extend in this paper.
Code2seq first converts an input snippet of source code into an AST, a tree-structured data representation given by a parser in a compiler.
After that, an encoder-decoder model is used to generate a function name from the AST.
We describe ASTs and the architecture of the base model below.

\subsection{Abstract Syntax Tree (AST)}

An AST is an intermediate representation used when source code is analyzed by a compiler.
The left part of Figure~\ref{ourmodel} shows an AST obtained from part of the source code in Figure~\ref{figinit}.
The leaves in the tree correspond to the strings that appear in the source code, and the non-terminal nodes are defined by the compiler.
The AST is obtained by using JavaParser.\footnote{https://github.com/javaparser/javaparser}

\subsection{Encoder}

As shown in Figure~\ref{ourmodel}, code2seq takes an obtained AST as an input.
It then extracts all possible shortest paths from a leaf to another leaf from the AST. 
Each path can be considered to be a sequence of nodes in the AST.
Next, it vectorizes two leaves and a sequence of non-terminal nodes on the shortest path. 
Each leaf $v$ is split into subwords and is converted into $\beta(v)$, 
the vector of the leaf that is defined as the sum of $e_w^{sub}$, the embedding vectors of the subwords $w$.

The sequence of non-terminal nodes is converted into a vector by using a bidirectional long short-term memory (LSTM)~\cite{Hochreiter:1997:LSM:1246443.1246450} based encoder:
$\overrightarrow{h_{t}}=\texttt{LSTM}(\overrightarrow{h}_{t-1},x_{t})$ and $\overleftarrow{h_{t}}=\texttt{LSTM}(\overleftarrow{h}_{t+1},x_{t})$,
where $x_{t}$ is the embedding of a non-terminal node.
The encoded vectors are concatenated as: $\gamma(v_1\cdots v_n)=[\overrightarrow{h}_n;\overleftarrow{h}_1]$, where `$;$' represents the concatenation of two vectors.
The vectors representing the two leaves and the vector representing the sequence of internal nodes are combined as follows:
\begin{equation}
q_{v_{0},v_{n+1}} = \notag tanh(W_{in}[\gamma(v_1\cdots v_n);\beta(v_0);\beta(v_{n+1})]),
\end{equation}
where $W_{in}$ is a matrix of a linear transformation of the concatenated vectors, and
$v_0$ and $v_{n+1}$ are the leaves of the beginning and end of the sequence, respectively.

\subsection{Decoder with Attention}
\label{decoder}
The decoder inherits the averaged vector of all possible paths between the leaf nodes in the AST as an initial state $s_{0}$.
To prevent the computational space from becoming too large, 
the maximum number of paths is set to 200; if there are more than 200 paths, 200 paths are randomly selected.
At each time step $t$, the decoder calculates the current hidden state $s_{t} = LSTM(s_{t-1}, y_{t-1})$, where $y_{t-1}$ is the embedding of the predicted subword in the previous time step. By using $s_{t}$, the decoder calculates the attention weights \cite{luong-etal-2015-effective} on the paths, each of which connects two leaf nodes. The weight on the $r$-th path is defined as follows:
\begin{align}
\label{path_vector}
a^t_r &=\frac{\exp(d_a^T\tanh(W_a[s_t;q_r]))}{\Sigma_{r'}\exp(d_a^T\tanh(W_a[s_t;q_{r'}]))},\\
p_{voc}(w) &= \sum_{i:w=w_i}{\delta^T_i softmax(W_l[\Sigma_{r}a^t_rq_r;s_t])},
\end{align}
where $q_r$ is the vector representation of the $r$-th path in the encoder, $W_a$ is a weight matrix for the linear transformation, and $d_a$ is a parameter vector.
Finally, the output layer calculates the label distribution at time step $t$ as $p_{voc}(w)$, where $W_l$ is a weight matrix.
$\delta_{i}$ is a one-hot vector, where only the $i$-th element is 1, and the others are 0. $w_i$ is the $i$-th subword in the vocabulary.

\section{Proposed Method}
We extend code2seq by adding a mechanism that generates low-frequency or out-of-vocabulary subwords. (Figure~\ref{ourmodel})
We propose two methods: the first one replaces the most frequent subword with a delexicalized placeholder; and the second one is a hierarchical copy mechanism.

\subsection{Placeholder for Most Frequent Subword}
In this method, we replace the most frequent subword in each input snippet of the source code with a placeholder {\ttfamily MFS} in the training data.
This idea is based on the observation that 31.26\% of function names in the training data include the most frequent subwords in the input snippets.
The existing methods have naively replaced all out-of-vocabulary subwords with a special tag, {\ttfamily UNK}.
We argue that such a strategy causes a lack of information for the tokens that should be included in the function name.
In comparison, our model knows which subwords are important even if they are out-of-vocabulary subwords, regarding the most frequent one as important. 
Thus, the model can more properly identify the important parts in the source code by assuming that a sequence containing the placeholder {\ttfamily MFS} is important.
When {\ttfamily MFS} appears in the output in the generation phase, we replace it with the original subword. 

\subsection{Hierarchical Copy Mechanism}
\label{Sec:HierachicalCopy}
\begin{figure}[t]
\centering
\resizebox{0.8\columnwidth}{!}{
\includegraphics{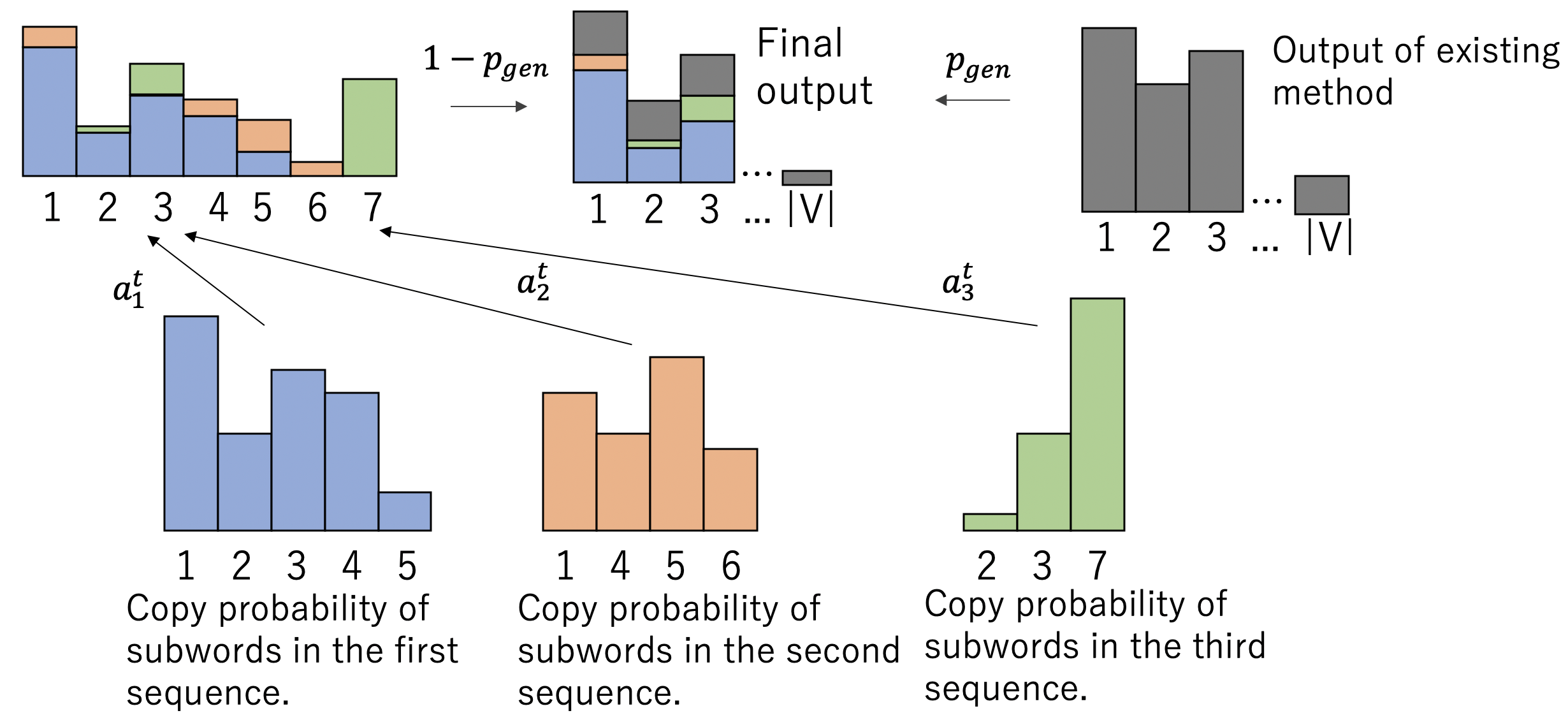}
}
\caption{Hierarchical copy mechanism. $|V|$ represents the vocabulary size.}\label{hattn} 
\end{figure}

In this section, we describe our proposed method with a hierarchical copy mechanism (Figure~\ref{hattn}). 
We observe that 71.42\% of function names in the training data include at least one of  the subwords in the input source code.
This led us to the idea of integrating a copy mechanism into code2seq.
The final probability of generating subword $w$ is the weighted sum of the probability $p_{voc}$ of generating the subword from the vocabulary and the probability $p_{copy}$ of copying the subword from the input:
\begin{align}
\label{Eq:CombCopyGen}
p(w)=p_{gen}p_{voc}(w) + (1- p_{gen})p_{copy}(w),
\end{align}
where $p_{gen}$ is the probability of selecting $p_{voc}$.

Conventional copy mechanisms cannot be directly applied to our task because their input is assumed to be a sequence of words, while the input in our setting is a tree or a set of paths containing tokens.
In particular, we propose a copy mechanism for copying subwords at leaf nodes.
In our model, to calculate the copying probability $p_{copy}(w)$, our hierarchical method combines two weights, the weight $a_r^t$ on the $r$-th path and the weight $b_{r,j}^t$ on the $j$-th subword on the $r$-th path, $w_{r,j}$:
\begin{align}
p_{copy}(w)=\sum_{r}\sum_{j:w=w_{r,j}}a_r^t b_{r,j}^t.
\label{eq1}
\end{align}
We use the conventional attention weights~\cite{luong-etal-2015-effective} for $a_i^t$, as described in Section 2.3; and $b_{i,j}^t$ is calculated as follows:
\begin{align}
    b_{r,j}^t&=\sum_{j} \delta_{j}^T softmax( h_{ctx}^T E_{r}^{sub}), \\
    h_{ctx}&=W_h a^t + W_s s_t + W_x e^{sub}_{y_{t-1}} + W_c g_t,\\
    g_t &=\sum_{k=0}^{t-1} a^k.
\end{align}
In the equation above, $E_{r}^{sub}$ is a matrix consisting of the embeddings of all subwords on the $r$-th path.
$\delta_{j}$ is a one-hot vector, where only the $j$-th element is 1, and the others are 0.
$a^t$ is a vector whose elements are the attention weights $a_r^t$ for each path $r$ at time step $t$, \fujisyo{and $e^{sub}_{y_{t-1}}$ is an embedding of the previous output subword.}
Thus, $g_{t}$ is a vector storing the sum of all attention weights at the previous time steps for every path.
$W_h$, $W_s$, $W_x$, $W_c$ are weight matrices, and $w_h'$, $w_s'$, $w_x'$, $w_c'$ are weight vectors for the linear transformation.

$p_{gen}$ is then calculated as follows:
\begin{align}
    p_{gen}&=sigmoid(h'_{ctx}),\label{eq2} \\
    h'_{ctx}&=w^{\prime~T}_h a^t + w^{\prime~T}_s s_t + w^{\prime~T}_x e^{sub}_{y_{t-1}} + w^{\prime~T}_c g_t.
\end{align}

\section{Experiments}
\subsection{Experimental Settings}

We evaluated our approaches on the following two datasets: Java-small and Java-large.\footnote{https://github.com/tech-Srl/code2seq\#datasets} 
Java-small consists of 691,974 functions  for training, 23,844 for development, and 57,088 for testing.
Java-large consists of 15,344,512 functions  for training, 320,866 for development, and 417,003 for testing. 
The models for comparison are as follows:
\begin{itemize}
\item \textbf{Code2seq}\\
We described the model in Section \ref{sec:code2seq}.
We reran the code\footnote{https://github.com/tech-Srl/code2seq} of \newcite{alon2018codeseq}.
\item \textbf{Copy}\\
This is a 2-layer LSTM-based pointer-generator model \cite{see-etal-2017-get}.
We experimented with OpenNMT-py\footnote{https://github.com/OpenNMT/OpenNMT-py} with 
the copy\_attn option.
\item \textbf{Pointer}\\
This is a variant of our hierarchical copy mechanism. Following the decoder of \newcite{fernandes2018structured},\footnote{They reported the state-of-the-art F1 scores on the Java small dataset. However, we could not reproduce their results with their code (https://github.com/CoderPat/structured-neural-summarization); other researchers also reported that they could not.(https://github.com/CoderPat/structured-neural-summarization/issues/25)} this model only points to tokens \cite{Vinyals:2015:PN:2969442.2969540} and does not generate any tokens. This model was prepared to verify the report of \newcite{fernandes2018structured} that a pointer-network works effectively and yields higher F1 scores than code2seq on the function naming task.
\end{itemize}

For training all the models, we used momentum-SGD~\cite{Qian:1999:MTG:307343.307376} as an optimizer. 
The batch size was set to 256, and the dimension of subword embeddings was 128. The dimension of the hidden layer in the encoder was set to 128, and that in the decoder was set to 320.
As a preprocessing, We split function and variable names in the source code into a sequence of subwords at the positons just before an uppercase character follows lowercase characters because programmers generally use camel case when writing code with Java. 
Long variable names were truncated to have at most 6 subwords.
We used only the paths that had less than 9 subwords.
We used TensorFlow to implement our models.

We used F1 as an evaluation metric, following \newcite{alon2018codeseq}, and added accuracy as another.

Furthermore, to correctly evaluate outputs with repeating tokens, we also used modified-F1 (F1**), calculated with the \textit{modified unigram precision} of \newcite{papineni-etal-2002-bleu} and \textit{unigram recall} of \newcite{lin-2004-rouge}.
F1** can prevent the models that repeatedly output subwords in the Gold function name from unreasonably obtaining high scores.

We calculated the above metrics on the basis of the number of subwords.
The accuracy measure was defined to be the number of correctly generated function names divided by the total number of test instances.
Here, we supposed an output is correct only if it is completely the same as the gold function name, while we calculated the other metrics by counting the overlap of subwords between generated function names and gold function names.
We trained and evaluated each model three times and computed the averaged scores.

\subsection{Results}

\begin{table}[t]
  \begin{minipage}[t]{.48\textwidth}
    \begin{tabular}{p{2em}p{6.0em}p{2em}p{2em}p{2em}}
    \toprule
    Corpus& Model & F1 & F1**& Acc\\
    \midrule
    Small&Code2seq$\ast$ & 43.02 & ~~~$-$ & ~~~$-$ \\ 
    &Code2seq &42.81 & 40.69 & 16.20 \\ 
    &Copy & 32.11& 31.94 &  16.33  \\
    &Pointer &\textbf{47.91}& 23.24& ~~5.49  \\ 
    \cmidrule(r){2-5}
    &Ours & 47.52& 45.34 &$\textbf{21.27}\dagger$ \\ 
    & w/o HierCopy &44.45 &42.90& 17.32  \\
    & w/o Replace & 47.08 &$\textbf{45.55}\dagger$& 19.13  \\
    \cmidrule(r){1-5}
    Large&Code2seq$\ast$& 59.19 &~~~$-$& ~~~$-$ \\
    &Code2seq &59.00 &58.16& 35.75 \\
    &Copy & 49.16& 48.97 &30.14  \\
    &Pointer&53.62& 27.83 &~~4.28  \\ 
    \cmidrule(r){2-5}
    &Ours & 58.96 &$\textbf{58.43}\dagger$ & 36.41 \\
    &w/o HierCopy& 58.69& 57.97 & 35.45 \\
    &w/o Replace& $\textbf{59.59}\dagger$ & 58.40  &$\textbf{36.61}\dagger$ \\
    \bottomrule
  \end{tabular}
   \caption{Evaluation results. `$\ast$' indicates the reported scores in the paper \cite{alon2018codeseq}. w/o indicates our model without the corresponding method. The highest score in each metric is shown in bold. $\dagger$ indicates that the difference from the best baseline was statistically significant with the paired bootstrap resampling method \cite{koehn-2004-statistical} ($p<0.001$).
}\label{fig10}
  \end{minipage}
  \hfill
  \begin{minipage}[t]{.48\textwidth}
    \begin{tabular}{p{2em}p{2em}p{6.5em}p{2em}p{2em}}
    \toprule
     Corpus & Target &Model&F1** &Acc\\
    \midrule
    Small&Low&Code2seq & ~~0.08 & ~~0.03 \\
    \cmidrule(r){3-5}
    &&w/o Replace& $\textbf{18.47}\dagger$ &$\textbf{10.69}\dagger$ \\
    \cmidrule(r){2-5}
    &Gen&Code2seq& 27.74  & 22.46 \\
    \cmidrule(r){3-5}
    &&w/o Replace&  $\textbf{30.40}\dagger$ &  $\textbf{24.20}\dagger$\\
     \cmidrule(r){1-5}
     Large&Low&Code2seq&  20.56 & 11.49\\
    \cmidrule(r){3-5}
    &~&Ours  & $\textbf{31.21}\dagger$&$\textbf{20.74}\dagger$ \\
    \cmidrule(r){2-5}
    &Gen&Code2seq&  40.19 & $\textbf{35.90}\dagger$\\
    \cmidrule(r){3-5}
    &&Ours&$\textbf{40.35}\dagger$& 35.65 \\
    \bottomrule
  \end{tabular}
  \caption{Additional results for different types of target subwords. Low means only words that appear with a probability of less than 0.0001\% in each corpus.
  \fujisyo{About 3\% of the function names contained one or more such words.}
  Gen means only words that do not appear in input source code.
  \fujisyo{Almost 8\% of the function names contained one or more such words.}
  Accuracy was calculated only with instances that contain at least one target subword.} \label{fig11}
  \end{minipage}
\end{table}

Table~\ref{fig10} shows the results.
On Java-small, our models scored higher than
code2seq in all the metrics. 
In particular, F1** increased by 2.21 points with the replacement strategy, 4.86 points with the model with our copy mechanism, and 4.65 points with the combination.
However, our best model had a lower F1 score than Pointer. This is consistent with the report of \newcite{fernandes2018structured}, indicating that function names commonly contain many subwords included in the given source code. 
The reason why F1** for Pointer was significantly lower than F1 is probably that it repeatedly outputs the same tokens.
The increase of the repetition may be caused by copying subwords from the small vocabulary included in the inputs.
These results suggest that the models for this task need to generate subwords not included in the given source code in order to correctly generate function names.
Moreover, the decoder part of seq2seq models is essentially the same as a unidirectional language model.
For that reason, the vanilla decoder requires a large amount of training data for generating various tokens in the output. The pointer network can help the decoder to generate various tokens without training token embeddings in the decoder side. Thus, the pointer network can work even with a small amount of training data.

In contrast, our models significantly outperformed Pointer in terms of F1** and accuracy. 
Our replacement strategy contributed little to F1** but significantly to accuracy. 
This is probably because our copy mechanism is effective in the decoding, whereas the replacement of the most frequent subwords helps to capture important information of the input in the encoder part. 
Thus, the combination can help both the encoder and decoder.

On Java-large, our models (the combination and with a copy mechanism) scored the highest in both F1** and accuracy among all the models. 
F1** increased by 0.27 points for the model with our copy mechanism, but the replacement strategy did not contribute at all.
This shows that replacing the most frequent subwords with a special token leads to ignoring their original meaning, that causes a disadvantage in a large corpus.
These results show the effectiveness of our hierarchical copy mechanism.

To check whether our best model can actually handle low-frequency or unknown subwords, we compared the best baseline and our models with the highest F1** score only on the subwords that appeared with a probability of less than 0.0001\% (Low in Table~\ref{fig11}).
On Java-small, code2seq could hardly handle the words with the probability less than 0.0001\%. 
In contrast, our method could output low-frequency words.
On Java-large, while code2seq could output some infrequent words, our method handled infrequent and out-of-vocabulary words better.

To investigate the importance of generation rather than copying, 
we examined the performances of the best baseline and our models with the highest F1** score only for subwords not included in the input (Gen in Table~\ref{fig11}).
On Java-small, our proposed method outperformed code2seq even in cases where we focus only on subwords not included in the input.
This is probably because the copy mechanism makes it easier to learn attentions with a small dataset.
On Java-large, our method outperformed code2seq in F**, but did not outperform it in accuracy.
It seems that our method emphasizes copying too much because function names tend to contain subwords in the input.
The scores for Pointer were always almost 0 on both datasets in the Gen setting because it cannot generate any subwords, even though it achieved the highest F1 score on Java-small. In contrast, comparing between Tables 1 and 2, 
the scores for our best model did not drop significantly even for the subwords not included in the input. 
These results suggest that the generation mechanism is necessary for the function naming task.

\subsection{Analysis}

\subsubsection{F** Scores for Subwords with Different Frequencies}

\begin{figure}[t]
\centering
\includegraphics[width=0.8\columnwidth]{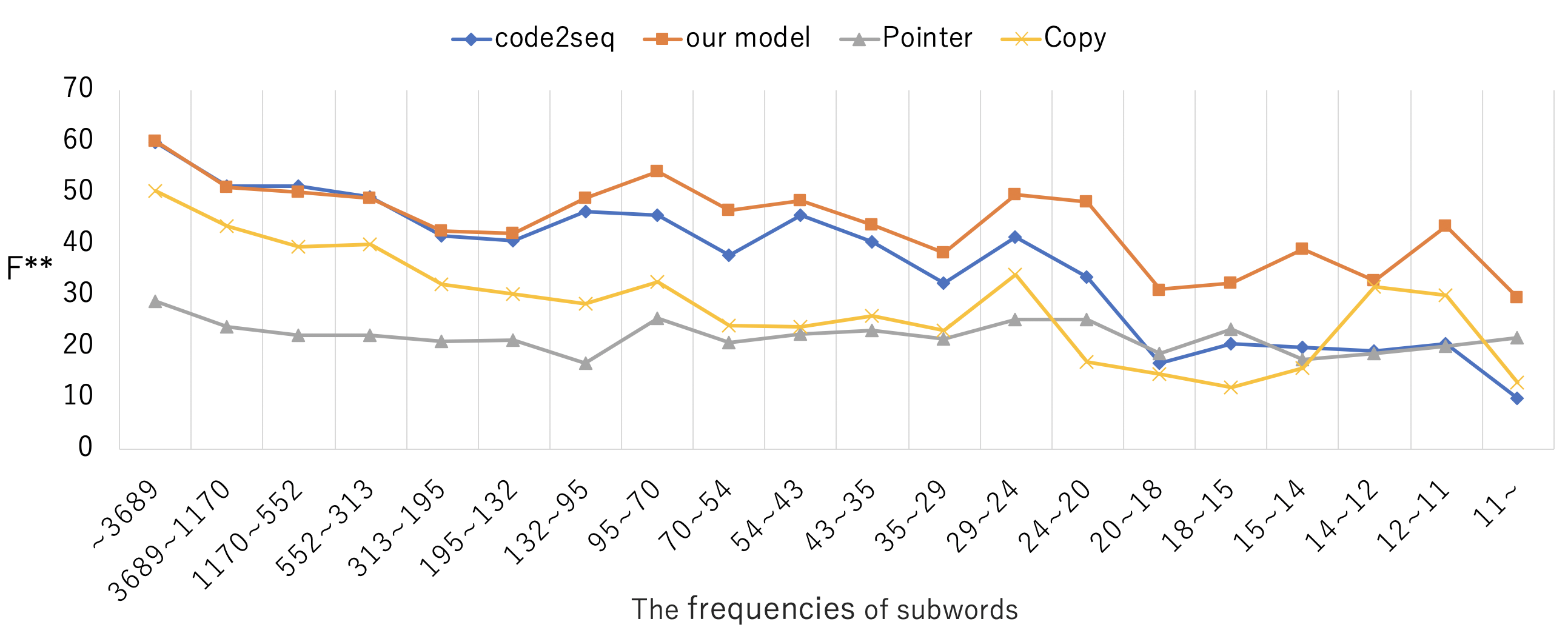}
\vspace{-1mm}
\caption{F** scores for subwords with various frequencies.}\label{graph} 
\end{figure}

We analyzed the relationship between F** scores of each model and the frequencies of subwords.
For this purpose, we first sorted subwords by their frequencies and after that, we split them into 20 classes equally.
We then calculated F** scores for the subwords in each class separately.

Figure~\ref{graph} shows the F** scores of each model for each class.
The leftmost class is the most frequent subwords, and the rightmost is the least frequent subwords.
The F** scores tend to decrease when the frequencies of subwords decrease.
This observation supports our assumption that infrequent subwords are difficult to predict.
Overall, the F** score for the leftmost class is almost the same as the result in Table~\ref{fig10} because the subwords in the class are the majority of the test data.

Code2seq and our model scored almost the same when frequencies of subwords are greater than 132.
On the other hand, if the frequencies of subwords are less than 132, our model achieved higher F** scores than code2seq.
Copy, which regards the source code as a sequence, scored lower than code2seq when the frequencies of subwords are greater than 14.
If the frequencies of subwords are less than 14, Copy achieved higher scores than code2seq.
These results indicate that even a vanilla copy mechanism can handle low-frequency subwords.
However, our model achieved higher F** scores than Copy.
These results indicate that our copy mechanism, which considers the abstract syntax tree, can handle low-frequency subwords better than the vanilla copy mechanism.
Pointer, which does not have the generation mechanism in the decoder, scored lower than the other methods in high-frequency subwords.
On the other hand, its F** scores for subwords whose frequencies are more than 11 and less than 20 were almost the same as the scores of code2seq, and the F** score of Pointer for subwords whose frequencies are less than 11 was significantly higher than the score of code2seq.
Thus, copying subwords is more useful than generating subwords for infrequent subwords.
Different from the other models, our model achieved the highest F** scores for subwords whose frequencies are less than 11.
This result indicates that the substitution of the most frequent subword is also useful for infrequent subwords.

From these results, we can further conclude that our proposed hierarchical copy mechanism can handle low-frequency subwords in this task,  compared with other baselines.

\subsubsection{Outputs of each model}
\begin{figure}[t]
\centering
\includegraphics[width=0.7\columnwidth]{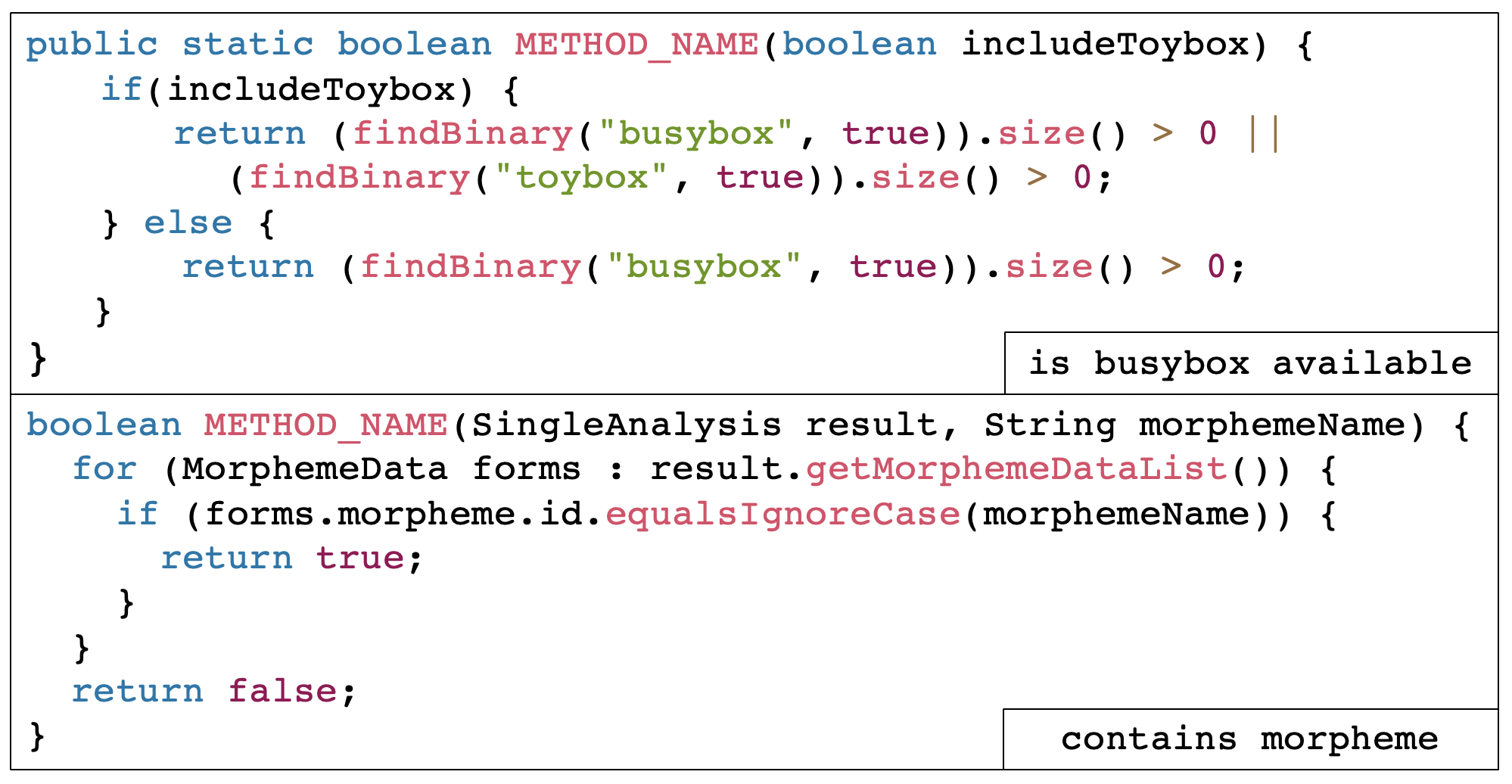}
\caption{Sample inputs.}\label{exp} 
\end{figure}

The top box of Figure~\ref{exp} shows a function that checks whether binary data with a predefined name such as {\ttfamily busybox} and {\ttfamily toybox} exist in the root directory.
Table~\ref{exp1out} lists generated method names from each model. 
Because {\ttfamily busybox} is a low-frequency word, code2seq did not generate it.
Copy generated {\ttfamily empty}, which does not appear in Gold.
Pointer copied {\ttfamily busybox} but it outputted the same word repeatedly.
In contrast, the output of our method is correct.

The bottom box of Figure~\ref{exp} is a function that checks whether the element of the first argument object has a morpheme of the second argument.
Table~\ref{exp2out} lists generated function names from each model.
Code2seq wrongly generated {\ttfamily data} instead of {\ttfamily morpheme}.
Copy generated {\ttfamily morpheme} correctly but the generated function name is not correct.
Pointer successfully copied {\ttfamily morpheme} but it also copied wrong words that are not related to the given function.
On the other hand, our method generated the function name which has a similar meaning to the one of Gold by replacing the most frequent subword in the given function.
Specifically, our method replaced {\ttfamily MFS} with {\ttfamily morpheme} because {\ttfamily morpheme} is the most frequent subword in this function.

As illustrated in the output from code2seq in Table~\ref{exp2out} that {\ttfamily morpheme} is generated as {\ttfamily data}, low-frequency words in a function might be replaced with more general-purpose words to explain how the function works.
However, if those words are replaced with the general words, many function names would become the same, and it would be difficult to differentiate between them.
Therefore, it is necessary to avoid using the general subwords such as {\ttfamily data} for generating function names. In that regard, our method is considered to be more practical because it can replace low-frequency words with MFS if they appear most frequently in the function.

\begin{table}[h]
  \begin{minipage}[t]{.45\textwidth}
    \begin{center}
      \begin{tabular}{cp{12.0em}}
            \toprule
            Method & Output\\
            \toprule
            Gold & is busybox available\\
            \midrule
            Code2seq & is available\\
            \midrule
            Copy & is empty \\
            \midrule
            Pointer & is busybox available busybox available busybox \\
            \midrule
            Our model & is busybox available\\
            \bottomrule
      \end{tabular}
    \end{center}
    \caption{Outputs for the top box of Figure~\ref{exp}}
    \label{exp1out}
  \end{minipage}
  \hfill
  \begin{minipage}[t]{.45\textwidth}
    \begin{center}
      \begin{tabular}{cp{12.0em}}
            \toprule
            Method & Output\\
            \toprule
            Gold & contains morpheme\\
            \midrule
            Code2seq & has data\\
            \midrule
            Copy & is morpheme \\
            \midrule
            Pointer & single analysis morpheme morpheme morpheme name morpheme \\
            \midrule
            Our model & has morpheme\\
            \bottomrule
      \end{tabular}
    \end{center}
    \caption{Outputs for the bottom box of Figure~\ref{exp}}
    \label{exp2out}
  \end{minipage}
\end{table}
\section{Related Work}
There have been a lot of research efforts on tasks where source code is the input.
\newcite{Hindle2012} and \newcite{DBLP:journals/corr/abs-1904-01873} constructed language models for the source code.
\newcite{Raychev:2015:PPP:2775051.2677009} proposed a method for outputting variable names in the source code.

\newcite{iyer-etal-2016-summarizing} proposed a model to summarize the behavior of functions in the source code.
\newcite{loyola-etal-2017-neural} proposed a method for generating descriptions of source code changes.

While these studies focus on source code as an input, their outputs are not function names.

As a method for representing source code, \newcite{pmlr-v37-allamanis15} converted a snippet of the source code into AST and proposed a method for generating a short description of the behavior of the snippet.
\newcite{allamanis2018learning} later proposed a method for detecting inappropriate variable names using AST.
We also used AST to represent the input snippet of source code while many other researches treat the source code as a sequence of tokens.

Regarding studies on function name generation, \newcite{DBLP:conf/sigsoft/AllamanisBBS15} proposed a method for generating function names using a stochastic language model that takes a sequence of tokens as an input, while we used a set of paths in AST.
\newcite{Alon:2018:GPR:3296979.3192412} formalized function name generation as a classification problem.
\newcite{Alon:2019:CLD:3302515.3290353} treated the same task as a sequence generation problem and proposed code2seq. 
In this paper, we proposed several extensions to code2seq. 

\newcite{Xu:2019:MNS:3294032.3294079} used a hierarchical attention network for function name generation.
In this model, the important information of the lower layer is passed to the upper layer by a recursive network.
Our model also took into account the hierarchical structure in our copy mechanism, as described in Section \ref{Sec:HierachicalCopy}.

We extended code2seq by adding the ability to copy subwords in the input source code.
The copy mechanism is a technique that copies subwords in the input to the output~\cite{gu-etal-2016-incorporating,gulcehre-etal-2016-pointing}.

Copy mechanisms have been shown to be effective in many tasks such as question-answering~\cite{he-etal-2017-generating}, document summarization~\cite{see-etal-2017-get}, headline generation~\cite{nallapati-etal-2016-abstractive} and question generation~\cite{zhao-etal-2018-paragraph}.
The existing copy mechanisms~\cite{nallapati-etal-2016-abstractive} presuppose a sequence of words as an input.
Although \newcite{Yang2018/05} and \newcite{hsu-etal-2018-unified} proposed a copy mechanism with hierarchical attention networks at word and sentence levels and \newcite{cohan-etal-2018-discourse} proposed a copy mechanism with hierarchical attention networks at word and section levels, they both assumed the input is a sequence of words, sentences, or sections.
Thus, their copy mechanisms cannot be directly applied to our setting because each input is assumed to be a set of paths in AST.
\newcite{fernandes2018structured} proposed a method for a function naming task using copy mechanisms.
They focused on extending the encoder, while we focused on extending the copy mechanism.
Our method used a hierarchy of copy layers rather than a single copy layer.

\section{Conclusion}

This paper dealt with the function name generation task.
We proposed two methods for including low-frequency or out-of-vocabulary subwords: replacing the most frequent subword in an input snippet of source code and with a hierarchical copy mechanism.
Our models outperformed the existing methods in terms of our modified F1 and accuracy.

Our proposed copy mechanism is applicable to tree-structured inputs such as discourse structures, cooking recipes, and social network services. Moreover, replacing the most frequent subword seems to be useful in tasks where the vocabulary is relatively small.

There remain two major issues to address.
The first is the need for better evaluation metrics.
We believe that this task requires a metric that can accept synonyms such as METEOR~\cite{banerjee-lavie-2005-meteor}.
However, some words that are considered synonymous in WordNet\footnote{https://wordnet.princeton.edu/} are used differently in the context of source code. 
For example, {\ttfamily increment} is an operation that increases the value of a variable by 1 in source code. It cannot be replaced with a word such as {\ttfamily increase}, even if they are synonymous with each other.
Therefore, we need an evaluation metric that takes into account the subtle difference between synonyms.

The second is to consider context in source code.
Our approach generates function names only from the information inside the function.
However, the behavior of other functions and the information on the objects handled by the function are important factors in generating the function name, because the function is called somewhere in the code.
Therefore, automatic generation of function names can be made more practical by considering the context in the source code.

\bibliographystyle{coling}
\bibliography{coling2020}

\end{document}